\documentclass[10pt]{article}
\usepackage[utf8]{inputenc}
\usepackage{cite}
\usepackage{amsmath,amssymb,amsfonts}
\usepackage[left=1.62cm,right=1.62cm,top=1.9cm]{geometry}
\usepackage{graphicx}
\usepackage{textcomp}
\usepackage{xcolor}
\usepackage{amsmath,amsfonts}
\usepackage{algorithm}
\usepackage{textcomp}
\usepackage{xcolor}
\usepackage[utf8]{inputenc}
\usepackage{listings}
\usepackage{float}
\usepackage{subfigure}
\usepackage{url}
\usepackage{multirow}
\usepackage{commath}
\usepackage{algpseudocode}
\usepackage{etoolbox}
\usepackage{varwidth}
\usepackage{tikz}
\usetikzlibrary{shapes,arrows}
\usepackage{colortbl}
\usepackage{xcolor}
\usepackage{booktabs}
\usepackage[colorlinks=true]{hyperref}
\usepackage{balance}
\definecolor{codegreen}{rgb}{0,0.6,0}
\definecolor{codegray}{rgb}{0.5,0.5,0.5}
\definecolor{codepurple}{rgb}{0.58,0,0.82}
\definecolor{backcolour}{rgb}{0.95,0.95,0.92}
\lstdefinestyle{mystyle}{
    backgroundcolor=\color{backcolour},   
    commentstyle=\color{codegreen},
    keywordstyle=\color{magenta},
    numberstyle=\tiny\color{codegray},
    stringstyle=\color{codepurple},
    basicstyle=\ttfamily\footnotesize,
    breakatwhitespace=false,         
    breaklines=true,                 
    captionpos=b,                    
    keepspaces=true,                 
    numbers=left,                    
    numbersep=5pt,                  
    showspaces=false,                
    showstringspaces=false,
    showtabs=false,                  
    tabsize=2
}

\def\BibTeX{{\rm B\kern-.05em{\sc i\kern-.025em b}\kern-.08em
    T\kern-.1667em\lower.7ex\hbox{E}\kern-.125emX}}

\begin{document}
\thispagestyle{plain}
\pagestyle{plain}

\title{BB-ML: Basic Block Performance Prediction using \\Machine Learning Techniques
}

\author{Hamdy Abdelkhalik${}^\ast$, Shamminuj Aktar${}^\ast$, Yehia Arafa${}^{\ast,\dagger}$, Atanu Barai${}^{\ast,\S}$, Gopinath Chennupati${}^\P$,\\ Nandakishore Santhi${}^\ddagger$, Nishant Panda${}^\ddagger$, Nirmal Prajapati${}^\ddagger$, Nazmul Haque Turja${}^{\ast,\S}$,\\ Stephan Eidenbenz${}^\ddagger$, Abdel-Hameed Badawy${}^{\ast,\ddagger}$}
\date{
    \small
    ${}^\ast$Klipsch School of ECE, New Mexico State University, Las Cruces, NM 80003, USA\\
    ${}^\ddagger$Los Alamos National Laboratory, Los Alamos, NM 87545, USA\\
    ${}^\dagger$Qualcomm Inc, USA  \hspace{2mm}      ${}^\S$Intel Corporation, USA    \hspace{2mm}    ${}^\P$Amazon Inc, USA\\
    \{enghamdy, saktar, yarafa, atanu, nhturja, badawy\}@nmsu.edu    \{nsanthi, nishpan, prajapati, eidenben\}@lanl.gov
}

\maketitle

\begin{abstract}
Recent years have seen the adoption of Machine Learning (ML) techniques to predict the performance of large-scale applications, mostly at a coarse level. In contrast, we propose to use ML techniques for performance prediction at a much finer granularity, namely at the Basic Block (BB) level, which are single entry, single exit code blocks that are used for analysis by the compilers to break down a large code into manageable pieces. We extrapolate the basic block execution counts of GPU applications and use them for predicting the performance for large input sizes from the counts of smaller input sizes.
In this work, we train a Poisson Neural Network (PNN) model using random input values as well as the lowest input values of the application to learn the relationship between inputs and basic block counts. Experimental results show that the model can accurately predict the basic block execution counts of $16$ GPU benchmarks. We achieve an accuracy of 93.5\% in extrapolating the basic block counts for large input sets when trained on smaller input sets and an accuracy of 97.7\% in predicting basic block counts on random instances.
In a case study, we apply the ML model to CUDA GPU benchmarks for performance prediction across a spectrum of applications. We use a variety of metrics for evaluation, including global memory requests and the active cycles of tensor cores, ALU, and FMA units. Results demonstrate the model's capability of predicting the performance of large datasets with an average error rate of 0.85\% and 0.17\% for global and shared memory requests, respectively. Additionally, to address the utilization of the main functional units in Ampere architecture GPUs, we calculate the active cycles for tensor cores, ALU, FMA, and FP64 units and achieve an average error of 2.3\% and 10.66\% for ALU and FMA units while the maximum observed error across all tested applications and units reaches 18.5\%.


\end{abstract}


\section{Introduction}
\label{sec:intro}

    


Graphics Processing Units (GPUs) have rapidly advanced to become the most common accelerators for high-performance computing. Modern GPUs are equipped with thousands of processors, capable of reaching up to 9.7 TFLOPS (trillion floating-point operations per second) for mixed-precision tasks~\cite{NVIDIAAm64:online}. Their highly parallel structure, exceptional floating-point computation capabilities, and memory parallelism make them ideal for a range of scientific and engineering applications in High-Performance Computing (HPC) environments. The most recent TOP500 list shows that the newest generation of supercomputers predominantly features GPU-accelerated systems~\cite{HomeTOP586:online}. This widespread adoption has sparked considerable research into the performance modeling and simulation of modern GPUs.

Modeling and simulation tools play a significant role in designing and evaluating new hardware features and understanding the limiting factors on application performance. Due to modern GPU complexity, improving its hardware performance has become more complex. Using ModSim(Modeling and Simulation) tools, GPU architects get insight into the impact of any changes in hardware design on application performance, area, and power consumption. Thus, these tools enable design space exploration. On the other hand, ModSim tools are also used extensively in hardware-software co-design to tune application performance. They help system designers choose the proper hardware for any specific workload. Additionally, application developers can tune their applications for a wide variety of GPUs without having access to the physical hardware. Scalable ModSim tools are necessary to allow quick and accurate design space exploration.

Performance analysis centered around the software concept of a basic block, defined as a single entry, single exit section of code, has recently been proposed for CPU architectures~\cite{ppt-ammp, abel2021BB, zhang-llvm-HPC, ppt-multicore}. The number of times a basic block is executed in a specific application is a good measure of the impact on the performance of that basic block for the whole application. A higher count leads to a higher impact. Thus, knowing the number of times (count) a BB executes is necessary for BB-level performance analysis. 

By characterizing the performance signature of the basic blocks and extracting the counts and other relevant information at a basic block level granularity, we can predict the performance of an application in finer granularity more efficiently and scalably. However, extracting the BB execution counts for large input sizes of an application can be very costly and become a scalability bottleneck, as it requires executing and/or profiling the application. Thus, addressing this inefficiency is necessary to achieve scalable performance prediction for ModSim tools, enabling hardware-software co-design. Therefore, BB count extrapolation enables scalable ModSim tools that use BB analysis in their performance prediction flow~\cite{ppt-ammp, ppt-multicore, WCET2013}. Furthermore, BB analysis is needed in compiler analysis passes~\cite{llvm-mca:online}, and this work could be leveraged for such purposes.

In this paper, we introduce BB-ML, which uses Deep Neural Networks (DNN) to predict basic block execution counts of GPU applications. First, we build a Poisson Neural Network (PNN). PNN is a probabilistic model that treats basic block counts as a Poisson random variable and leverages Deep Neural Networks to learn to predict basic block counts for GPU applications. 
It also can extrapolate basic block counts for larger application input configurations (sizes). To train the model, we use labeled data where the input parameters of applications are used as input features for the neural networks, while the BB counts are used as the output of the networks. Results show that our model can accurately extrapolate basic block counts for large input configurations and random input values of the applications.

The contributions of the paper are:

\begin{itemize}
    \item To our knowledge, this is the first work to predict BB counts for GPU applications.
    \item We introduce Poisson Neural Network (PNN), a probabilistic DNN model that treats BB counts as Poisson distributions to capture the relationship between input sets of a program and BB counts.
    
    \item We demonstrate that the trained model showed competitive accuracy on the set of benchmarks used. The model achieved 93.5\% accuracy for extrapolating basic block counts from higher input configurations and 97.7\% accuracy for predicting block counts on random instances.
    \item We applied our model on a study case for validation.
   
\end{itemize}

The rest of the paper is organized as follows: Section~\ref{sec:related-works} summarizes the relevant related work and sets this work apart from the relevant related work. Section~\ref{sec:background} covers the necessary background and shows a control flow graph with basic blocks of a small sample program. Section~\ref{sec:methodology} overviews the basic block trace extraction, the two neural network architectures, and their training. Section~\ref{sec:exp} goes over the experimental results of our NN models. It compares the two models and shows detailed results.  Finally, Section~\ref{sec:conclusion} concludes the paper.

\section{Related Works}
\label{sec:related-works}

BB-level performance prediction attracted much research in CPU~\cite{ppt-ammp,ppt-sasmm} and GPU~\cite{WCET2013} domains. Tools such as llvm-mca~\cite{llvm-mca:online} from LLVM or IACA~\cite{iaca:online} from Intel perform BB-level program analysis. This section shows some of the work centered around BBs, predicting their count, and throughput, or leveraging the BB  information to predict the whole program execution time.

In the CPU domain, Mendis~\textit{et al.}~\cite{mendis2019ithemal} proposed Ithemal, a tool for predicting the throughput of basic blocks in applications. The authors used a regression-based model alongside with a Long Short-Term Memory (LSTM) recurrent neural network to predict how many times the BB gets executed. Their model takes the BB assembly instructions as input and provides the basic block throughput as output. In the same spirit, Abel~\textit{et al.}~\cite{abel2021BB} proposed a parametric pipeline model and basic block throughput predictor for Intel processors. Their model also addresses some limitations in the Ithemal model. Other works, such as PPT-AMMP~\cite{ppt-ammp} proposed by Chennupati~\textit{et al.},  demonstrated a regression-based technique to predict the BB counts of CPU applications for large input sets. They used the counts to predict the run-time of serial applications on a single-core CPU.

In the GPU domain, Betts~\textit{et al.}~\cite{WCET2013} developed a model that uses basic block (BB) execution information, such as execution traces and time, to predict an application's worst-case execution time.  They extracted BB traces using a cycle-accurate simulator, which is notably slow; such simulations can take weeks or even months for extensive input sizes~\cite{arafa2021hybrid}. Another limitation is the necessity to extract traces anew with each input change, which is not a daunting task on cycle-accurate simulators.

To our knowledge, no prior work exists that predicts the BB executions throughput for GPU architectures. Adopting this work in the future can reduce the time and space of trace extraction for performance prediction models~\cite{arafa2021hybrid, WCET2013}. Thus, enabling scalable predictions that can train the model once per application on small input sizes and reconstruct a trace for higher input sizes.

\begin{figure}[b!]
    \centering
    \includegraphics[width=.3\linewidth]{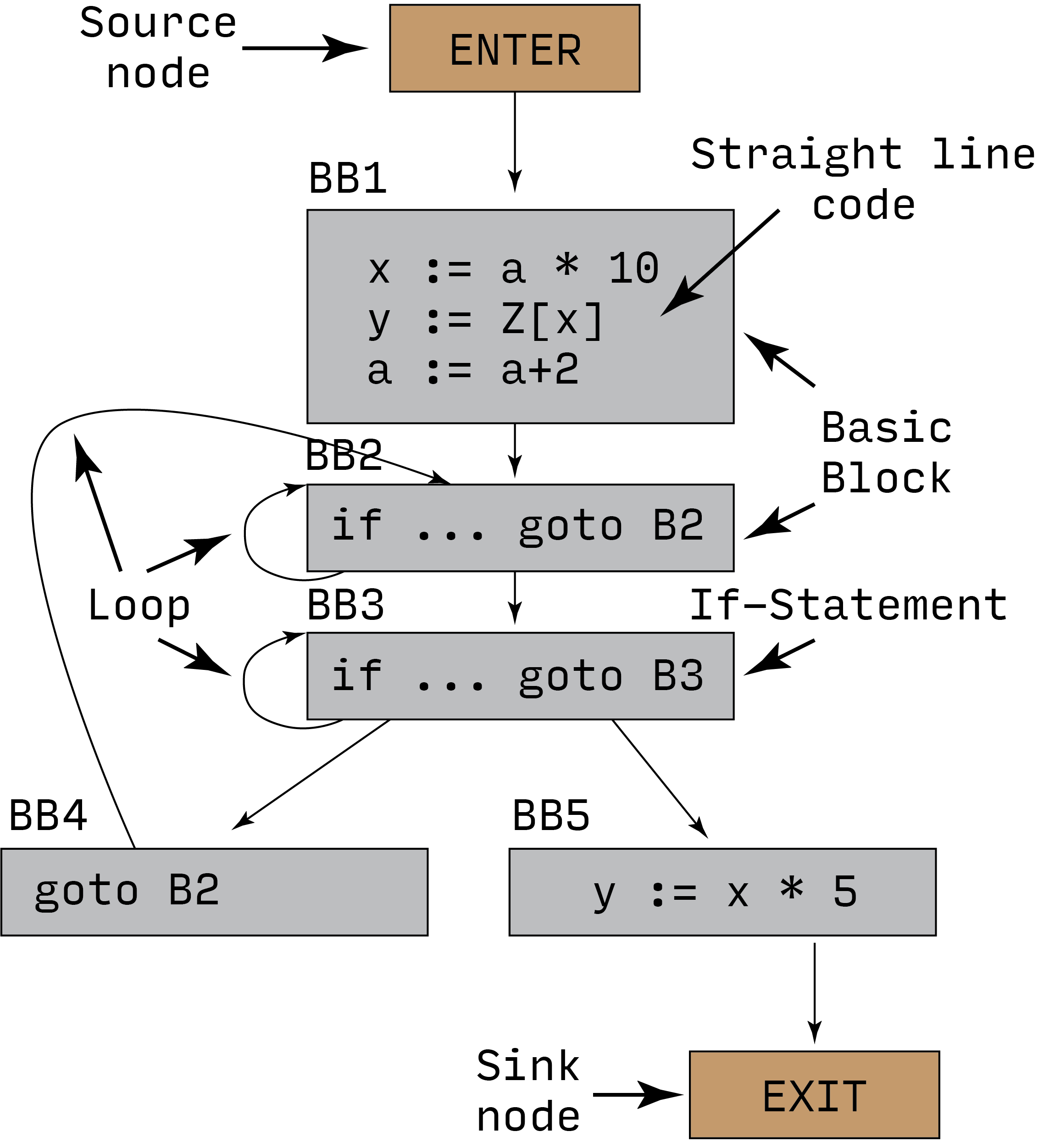}
        \caption{A control flow graph with five basic blocks . Each BB has a single entry and a single exit point. BB2, BB3, and BB4 represent a loop.
        }
        \label{fig:basic-block}
\end{figure}

\begin{figure}
	\centering
	\includegraphics[width=.25\linewidth]{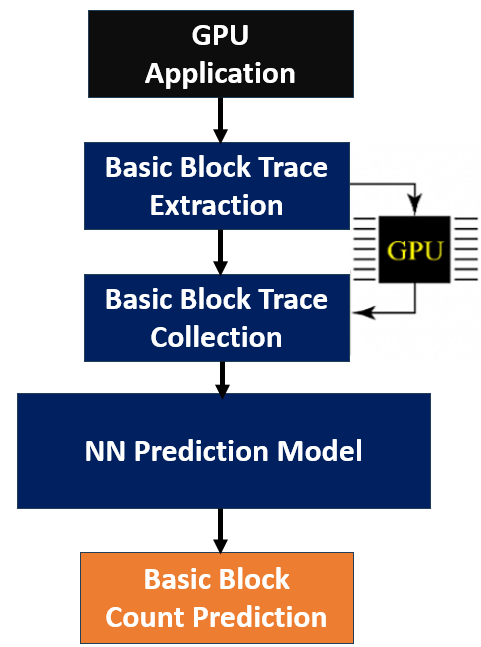}
    \caption{Overview of BB-ML}
    \label{fig:overview_flow}
    \vspace{-3mm}%
\end{figure}

\section{Background}
\label{sec:background}


\subsection{Basic Blocks}
A basic block represents a straight code sequence with no branches. In other words, a BB is a linear code sequence with a single entry and a single exit point with no branches in between. In a program Control Flow Graph (CFG), if a program execution enters a BB, all the instructions in that BB execute until a branch instruction is encountered. Generally, the compiler breaks down the program into multiple BBs during analysis.


Figure~\ref{fig:basic-block} shows the control flow graph of an example program. There are five BBs in this program. The BBs \emph{BB1 ... BBn} represent the vertices or the nodes in the CFG. A directed edge (\emph{BBi} \textrightarrow \emph{BBj}) exists if program control flows from \emph{BBi} to \emph{BBj}. Special nodes \textbf{Enter} and \textbf{Exit} are the \emph{source} and the \emph{sink} of the graph. A loop is composed of multiple basic blocks. For example, \emph{BB2}, \emph{BB3}, and \emph{BB4}  represent a loop. Conditional statements like if statements represent their own basic blocks, e.g., \emph{BB2} and \emph{BB3}.

\subsection{GPU Execution Hierarchy}
In the CUDA programming language, a group of threads is called a CUDA block. CUDA blocks are grouped into a grid. A GPU program or kernel is a function executed concurrently as a grid of thread blocks. GPU programs utilize data, thread, and task parallelisms. From the software perspective, GPUs have three execution levels: block-level, warp-level, and thread-level, to achieve such parallelisms. Additional information can be found in the CUDA programming model~\cite{Programm59:online}.

\subsection{Machine Learning Techniques}
Researchers from diverse domains widely employ machine learning for solving complex problems. Machine learning utilizes mathematical models and data analysis to learn information directly from data. Probabilistic models assist machine learning by examining the underlying distribution of datasets and learning the pattern from inputs and outputs. Regression is one of the machine learning techniques from supervised learning where the model learns from labeled samples.Regression methods are used to predict numerical outcomes based on insights gained from a dataset with predefined labels. The most commonly used regression techniques are linear regression, logistic regression, and polynomial regression~\cite{freund2006regression}.

Deep Neural Networks, inspired by human brain information processing, use hidden layers to store and find the relationship between the inputs and the outputs. A neural network consists of input, hidden, and output layers, where information is passed through the layers and updated using feedback mechanisms. The activation function, learning mechanism, and loss function are used in the neural network to optimize the training process. An artificial neural network composed of multiple hidden layers is a Deep Neural Network (DNN). DNNs are efficient in modeling complex non-linear real-world problems. Some commonly used DNN frameworks are TensorFlow~\cite{abadi2016tensorflow}, PyTorch~\cite{paszke2019pytorch}, and Keras~\cite{gulli2017deep}.

\section{Methodology} 
\label{sec:methodology}

This section goes over the various steps needed for BB count prediction. We first explain the BB trace extraction. Then, we detail the architecture of the Neural Network models, and, finally, the training of the models.

BB-ML can utilize any NVIDIA GPU, given its configuration parameters. We are limited to NVIDIA GPU since we use their tools for BB count profiling. The methodology can be extended to other GPUs as long as we can collect the necessary traces. Figure~\ref{fig:overview_flow} shows a high-level overview of BB-ML. It comprises various components that feed off of each other to predict the BB counts eventually. First, the \emph{BB Trace Extraction Tool} extracts BB  traces by running the application on a physical  GPU. Then, the \emph{collected BB traces} are fed to the two Neural Network models to get the \emph{BB Count Predictions}.

\begin{algorithm}[b!]
\caption{Trace Extraction Algorithm}
\label{alg:cap}
\begin{algorithmic}[1]
\State $ID\_BB = 0$     \hfill //Basic Block ID
\State $ID\_ker = 0$    \hfill //GPU Kernel ID
\State $BB\_counts\_array [$$1000$]\    
\State $kernel\_IDs\_array [$$1000$]\  
\Procedure{instrumentation\_function}{}
\While{$kernel\_found = 1$}
\While{$BB\_found = 1$}
\State $Send(\ $ID\_BB  )\
\State $Send(\ $ID\_ker )\
\State $ID\_BB \gets ID\_BB + 1$ 

\EndWhile
\State $ID\_ker \gets ID\_ker + 1$ 

\EndWhile
\EndProcedure
\Procedure{receive\_function}{}
\While{$data\_available = 1$}
\State $X \gets $Read(\ )\
\State $BB\_counts [\ X.ID\_BB]\  \gets BB\_counts [\ X.ID\_BB]\ + 1$
\State $kernel\_IDs [\ X.ID\_BB]\  \gets   [\ X.ID\_ker]\ $

\EndWhile
\EndProcedure
\end{algorithmic}
\end{algorithm}

\subsection{BB Trace Extraction}
\label{sec:data_collection}
The first step of the model is extracting the BB execution traces. To extract the BB trace, we extended NVBit~\cite{villa2019NVBit} to get the trace dynamically. NVBit is a dynamic binary instrumentation tool from Nvidia Research that can implement profilers, perform error checking, and collect various traces, including those of memory, instructions, and BBs. In our work, we adapted the NVBit tool to dynamically collect the BB counts per kernel for GPU applications.

We extract an application-dependent BB trace. Thus, we can use any Nvidia GPU. 
We extract the different application traces using several GPUs; a Tesla A100 Ampere GPU,  
a Titan V Volta GPU, and a Tesla V100 Volta GPU. 

Algorithm~\ref{alg:cap} shows a high-level overview of the logic of extracting the BB trace using NVBit. The first step for the algorithm is to instantiate and initialize the necessary variables (BB ID, Kernel ID, and the arrays holding the BB count for each BB and the array of kernel IDs) as shown in lines [1-4]. We set the array sizes to 1000 for illustration purposes, but the array size varies by the number of BBs for each kernel and the number of kernels in a GPU application. NVBit assigns an automatic ID for each BB and kernel, but these values are very large and not in order. For simplicity, we use the ID\_BB and ID\_ker variables in lines [1,2] to assign new in-order IDs for the BBs and the kernels. In lines [5-14], we go over all the kernels and the basic blocks in the application and inject a function after each BB to collect the IDs of the BB and the kernels. In lines [15-21], we use another function to receive the IDs dynamically and calculate the BB counts. In line 19, we store the kernel ID in the same index as the BB count because we need to associate the specific kernel that each BB belongs to.

We follow the methodology suggested in the NVBit tool to profile the applications at run-time. 
We wrote a script that automates the BB traces automatically while changing the input at each invocation. We collect the BB counts per thread block per kernel per application and feed it to the ML model afterward. 

\subsection{Deep Neural Network Model Architecture}
\label{sec:model-architecture}

This section describes the Poisson Neural Network Model (PNN), a probabilistic model that treats the underlying distribution of BB execution counts as a Poisson probability distribution. The task of predicting counts is often modeled as a regression problem in various domains~\cite{yan2011citation, iqbal2021covid, nevendra2019software}. 

\begin{figure}
	\centering
	\includegraphics[width=0.45\linewidth]{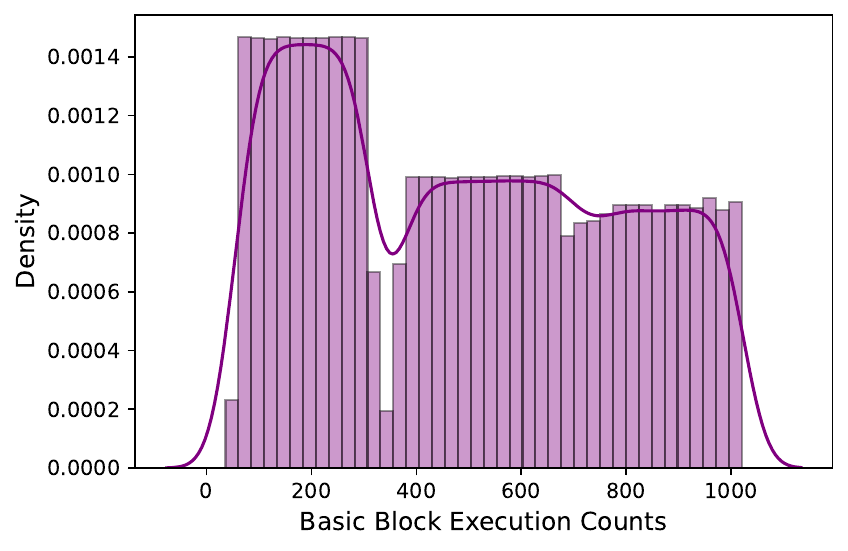}
    \caption{Kernel density estimation plot of $19^{th}$ basic block of bicg benchmark from Table~\ref{dataset} showing the underlying probability distribution of counts. The probability of each basic block count can be computed by finding the area under the curve on the x-axis.
	}
    \label{fig:count-dist}
    \vspace{-3mm}
\end{figure}

\begin{table*}[t]
\caption{Benchmark applications used from Rodinia~\cite{che2009rodinia} (indicated with $\ddagger$) and Polybench~\cite{grauer2012auto} (indicated with $\ast$) benchmark suites in this work. The table lists the total number of BB samples collected, input parameters, and BBs for each application. 
}
\centering
 \resizebox{0.99\columnwidth}{!}{%
	\begin{tabular}{@{}@{}c l c c c l@{}@{}}
		\toprule \textbf{Benchmark} & \textbf{Description} & \textbf{\# of BB Samples} &  \textbf{\# of Inputs} & \textbf{\# of Basic Blocks} & \textbf{Domain} \\
		 \toprule 
		 ${}^\ast$2mm & 2 Matrix Multiplications & 11505 & 4 & 21 & Linear Algebra\\
		 ${}^\ast$atax & Matrix Transpose and Vector Multiply & 25865 & 2 & 21& Linear Algebra\\
		 ${}^\ast$bicg & BiCGStab Linear Solver & 27281& 2 & 21& Linear Algebra\\
		 ${}^\ast$covariance & Covariance Computation & 9595 & 2 & 54 & Data Mining\\
		 ${}^\ast$correlation & Correlation Computation & 5020 & 2 & 135 & Data Mining\\
		 ${}^\ast$doitgen & Multiresolution Analysis Kernel & 6009 & 3 & 13 & Linear Algebra\\
		 ${}^\ast$gemm & Matrix Multiplications & 81219& 3 & 10& Linear Algebra\\
		 ${}^\ast$gesummv & Scalar, Vector-Matrix Multiply & 9998 & 1 & 10 & Linear Algebra\\
		 ${}^\ast$lu & LU Decomposition & 4347 & 1 & 31 & Linear Algebra\\
		 ${}^\ast$gramschmit & Gram-Schmidt Decomposition & 3117 & 1 & 83 & Linear Algebra\\
		 ${}^\ast$syrk & Symmetric Rank-k Operations & 9900& 2 & 15 & Linear Algebra\\
		 ${}^\ast$mvt & Matrix-Vector Product Transpose & 8189 & 1 & 21 & Linear Algebra\\
		 ${}^\ddagger$gaussian & Gaussian Elimination & 2070 & 1 & 36 & Linear Algebra\\
		 ${}^\ddagger$lud & LU Decomposition & 2052 & 1 & 11 & Linear Algebra\\
		 ${}^\ddagger$nw & Needleman-Wunsch & 3357 & 1 & 21 & Bioinformatics \\
		 ${}^\ddagger$pathfinder & Path-Finder& 65171 & 3 & 6 & Dynamic Programming\\
		\hline
	
	\end{tabular}
 }
	\label{dataset}
\end{table*}

A Poisson probability distribution models the occurrence of an event within a fixed period of time. It is commonly used to model a system where the target variable is a discrete count variable. BB execution counts are always positive and discrete in nature. We analyze the distribution of BB counts and their relationship to the input parameters. Figure~\ref{fig:count-dist} presents a kernel density estimation (KDE) plot of counts for one BB of the `bicg' dataset from Table~\ref{dataset}. The X-axis represents the BBs counts, and the Y-axis shows the probability density function for the kernel density estimation. The density plot shows the underlying distribution of BB counts, which is Poisson-like, where each count is independent. From the distribution, we find that BB counts can be modeled as a Poisson random variable with mean $\lambda$. We propose a machine learning model that can learn mapping $(f)$ from input parameters $(x)$ to predict BB count. Suppose $x$ is our input parameter vector, and $Y$ is the BB count. In that case, we model the BB count as a Poisson random variable with mean $\lambda$, where $\lambda$ only depends on input parameter $x$. Thus, the probability that $Y$ has count $j$ is given by:
\begin{equation}
    P(Y=j) = \dfrac{e^{-\lambda(x)}\lambda(x)^j}{j!}
\end{equation}
The Poisson Neural Network (PNN) uses a Poisson loss function to minimize the loss and maximize the likelihood of the data. PyTorch's Poisson negative log likelihood $(nn.Poisson NLLLoss)$ loss function is used to compute the loss in the training process where we choose the $log\_input$ parameter to be $False$~\cite{poisson-loss}.
The loss is a scalar value that is computed from input and target as,
\begin{equation}
    loss(input,target) = input - target * log(input + eps)
\end{equation}
\noindent where $eps$ is a small value which is added to avoid computation of $log(0)$. We implemented the PNN model using PyTorch. The input layer neurons are fully connected to the hidden layer, and the hidden layer is also fully connected to the output layer. The hidden layer's size is ten, and the output layer is 1, where the input layer size depends on the number of inputs of the application. We used the $tanh$ activation function to transform the weighted sum of the input into an output. $Adam$ optimizer is used in PNN to update weights and the learning rate in the training stage. We tune the hyperparameters (number of epochs, batch size, and learning rate) to optimize the training process. Rigorous tuning shows that the network gets minimal loss for specific values of the hyperparameters. Therefore, we train the models for 300 epochs using a batch size of 10 and choose the learning rate to be $10^{-4}$. 


\begin{table*}[t!]
\centering
    \caption{Average MSE and correlation for basic block counts prediction on random input instances for the benchmarks listed in Table~\ref{dataset}.\\}
   
    \resizebox{0.99\columnwidth}{!}{%
	\begin{tabular}{|c c c c||c c c c|}
        \hline
		\textbf{Benchmark} & \textbf{Avg.~MSE} & \textbf{Pearson Corr.} & \textbf{Spearman Corr.} & \textbf{Benchmark} & \textbf{Avg.~MSE} & \textbf{Pearson Corr.} & \textbf{Spearman Corr.} \\
	   \hline
	   2mm & 0.015 & 0.999 & 0.999 & gesummv & 0.001 & 0.999 & 1.000 \\
        \hline
	   atax & 0.003 & 0.999 & 0.999 & mvt & 0.024 & 0.999  & 0.999 \\
	   \hline
	   bicg & 0.008 & 0.999  & 0.999 & syrk & 0.002 & 0.999  & 0.999 \\
	   \hline
	   gemm & 0.002 & 0.999  & 0.999 &  gramschmit & 0.002 & 0.999 & 1.000 \\
	   \hline
	   gaussian & 0.080 & 0.939 & 0.928 & lud & 0.048 & 0.958  & 0.973 \\
	   \hline
	   pathfinder & 0.001 & 0.999 & 0.998 & nw & 0.012 & 0.999 & 0.999 \\
        \hline
	   lu & 0.177  & 0.850  & 0.929  & correlation & 0.002 & 0.999 & 0.997\\
	   \hline
	   covariance & 0.001 & 0.999 & 0.999 & dotigen & 0.001 & 0.999 & 1.00 \\
		\hline
	\end{tabular}
	}
	\label{random-result}
	
\end{table*}
\begin{table*}[t!]
\centering
 
    \caption{Average MSE and correlation for basic block counts extrapolation on high input instances for the benchmarks listed in Table~\ref{dataset}.\\}
   
    \resizebox{0.99\columnwidth}{!}{%
	\begin{tabular}{|c c c c||c c c c|}
        \hline
		\textbf{Benchmark} & \textbf{Avg.~MSE} & \textbf{Pearson Corr.} & \textbf{Spearman Corr.} & \textbf{Benchmark} & \textbf{Avg.~MSE} & \textbf{Pearson Corr.} & \textbf{Spearman Corr.} \\
	   \hline
	   2mm & 0.019 & 0.999 & 0.999 & gesummv & 0.030 & 0.999 & 0.999 \\
        \hline
	   atax & 0.036 & 0.998 & 0.999 & mvt & 0.030 & 0.998  & 0.999 \\
	   \hline
	   bicg & 0.017 & 0.999  & 0.999 & syrk & 0.022 & 0.994  & 0.999 \\
	   \hline
	   gemm & 0.026 & 0.999  & 0.999 &  gramschmit & 0.054 & 0.996 & 0.999 \\
	   \hline
	   gaussian & 0.159 & 0.757 & 0.797 & lud & 0.178 & 0.907  & 0.922 \\
	   \hline
	   pathfinder & 0.094 & 0.995 & 0.997 & nw & 0.115 & 0.980  & 0.999 \\
        \hline
	   lu & 0.208  & 0.648  & 0.621  & correlation & 0.087 & 0.983 & 0.984\\
	   \hline
	   covariance & 0.039 & 0.998 & 0.999 & dotigen & 0.001 & 0.999 & 1.00 \\
		\hline
	\end{tabular}
	}
	\label{exp-result}
\end{table*}
\subsection{Training the Neural Network}
\label{train-method}
We have three sets of training and testing data generated from the same original collected data. We split the data into high-low, mixed high-low, and random data. Here, we explain how the data is split.
First, for the high-low data, we assign the lower 70\% of the input range of the input samples to be the training set and the remaining 30\% of the input range, which represents the higher range input samples to be the test set. There are some input values with a combination of low and high values. We first train our models without these mixed input values and calculate accuracy. To examine the effect of mixed input values, we also train the models with a training dataset that includes the mixed range of input samples (mixed high-low data). Lastly, we randomly assign 70\% of the samples to a training set and the rest 30\% to the testing set to measure the overall accuracy of our model.

Prediction and extrapolation on the PNN model from Section~\ref{sec:model-architecture} generate scalar values of basic block counts. To compare the predicted and actual basic block counts, we use a normalized error metric, Mean Squared Error (MSE), defined as follows,

\begin{equation}
    MSE = \frac{1}{n} \sum_{i=1}^{n} (Y_i - \hat{Y_i})^2
\end{equation}

\noindent where $n$ is the number of samples, $Y_i$ is the predicted counts, and $\hat{Y_i}$ is the actual basic block count. 

We also compute the Pearson correlation coefficient~\cite{pearson-corr}, and Spearman correlation coefficient~\cite{spearman-corr} between predicted values and actual values of basic block execution counts for each benchmark. Pearson correlation measures the strength of the linear relationship between predicted and measured counts. Spearman correlation assesses whether there is a linear relationship between variables equal to the Pearson correlation between the rank values of those two variables.

\begin{figure}
\vspace{-3mm}
    \includegraphics[width=.25\textwidth,height =.2\linewidth]{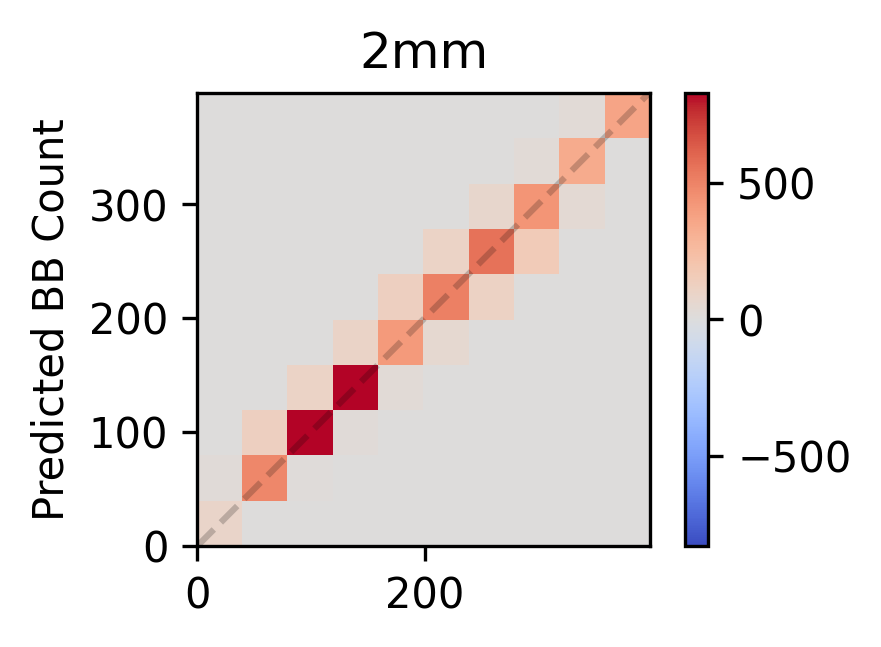}\hfill
    \includegraphics[width=.25\textwidth,height =.2\linewidth]{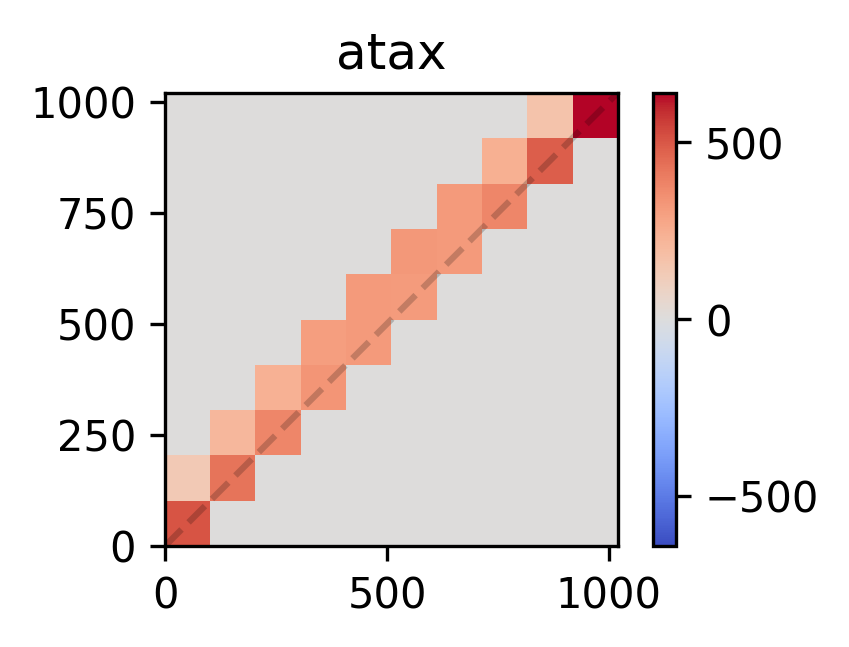}\hfill
    \includegraphics[width=.25\textwidth,height =.2\linewidth]{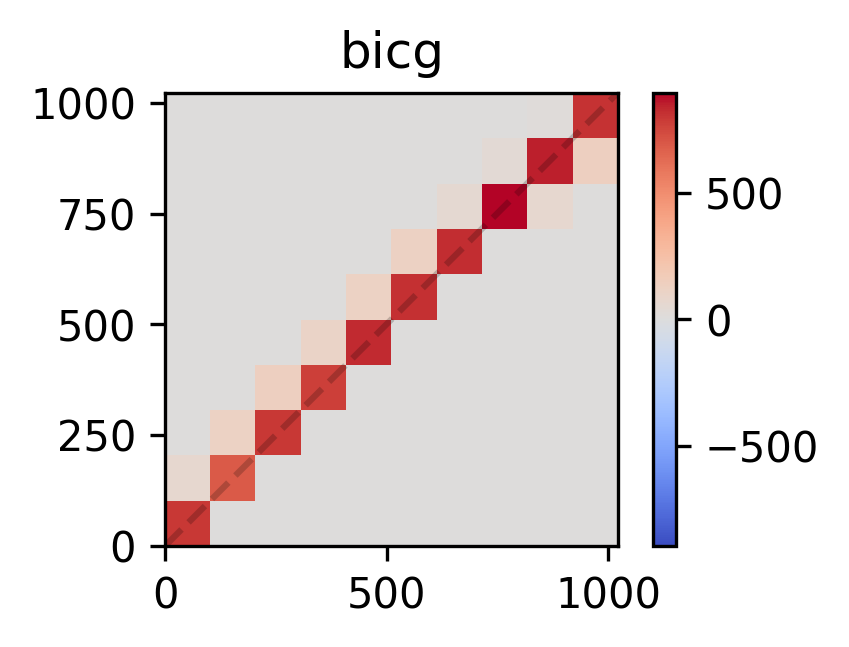}\hfill
    \includegraphics[width=.24\textwidth,height =.2\linewidth]{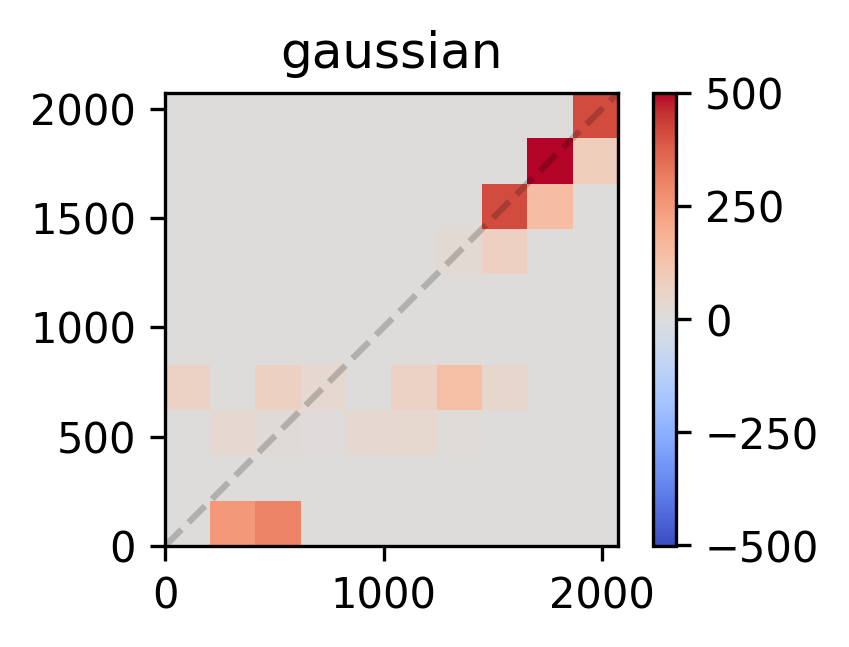}\hfill
    \includegraphics[width=.25\textwidth,height = .2\linewidth]{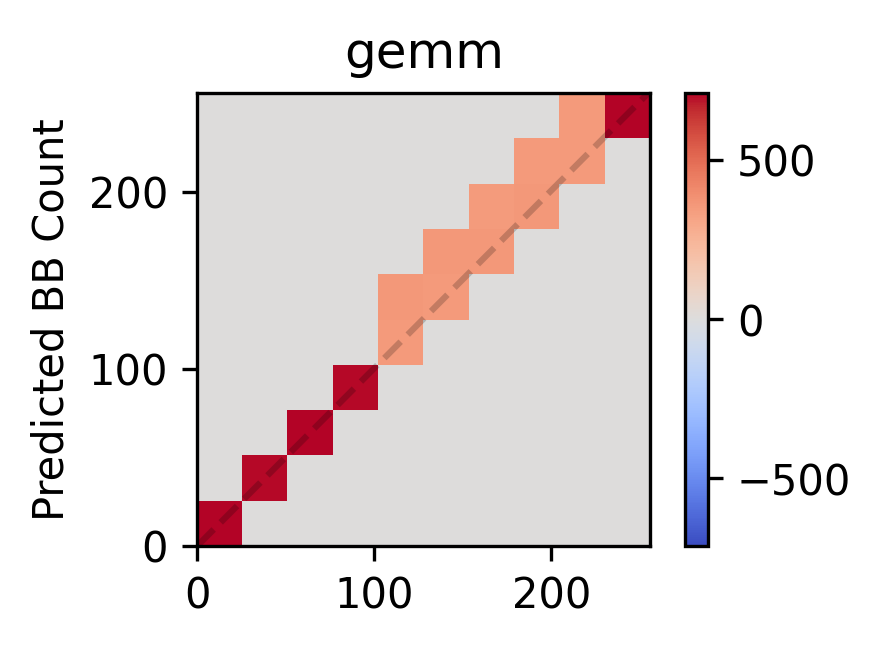}\hfill
    \includegraphics[width=.25\textwidth,height = .2\linewidth]{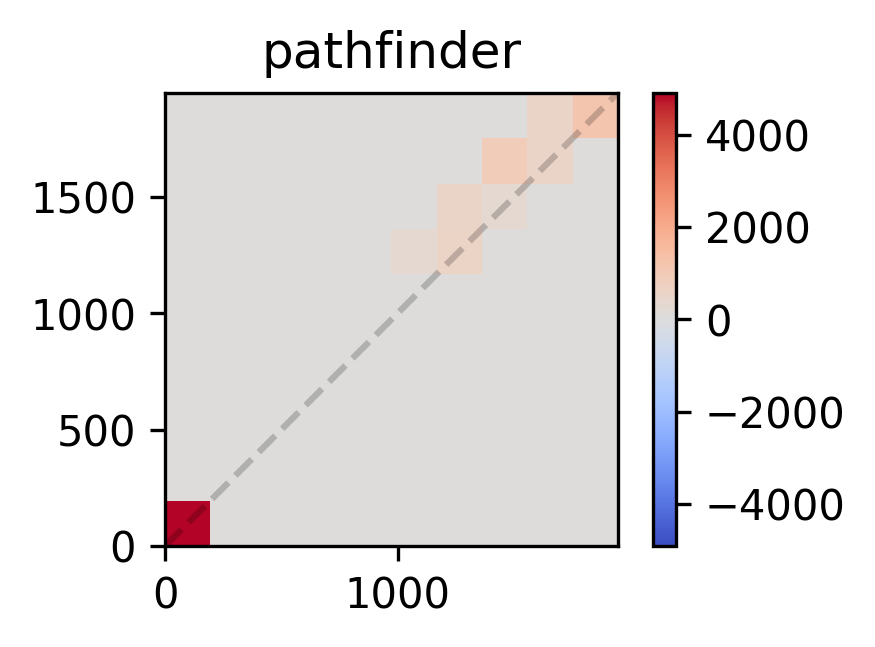}\hfill
     \includegraphics[width=.25\textwidth,height = .2\linewidth]{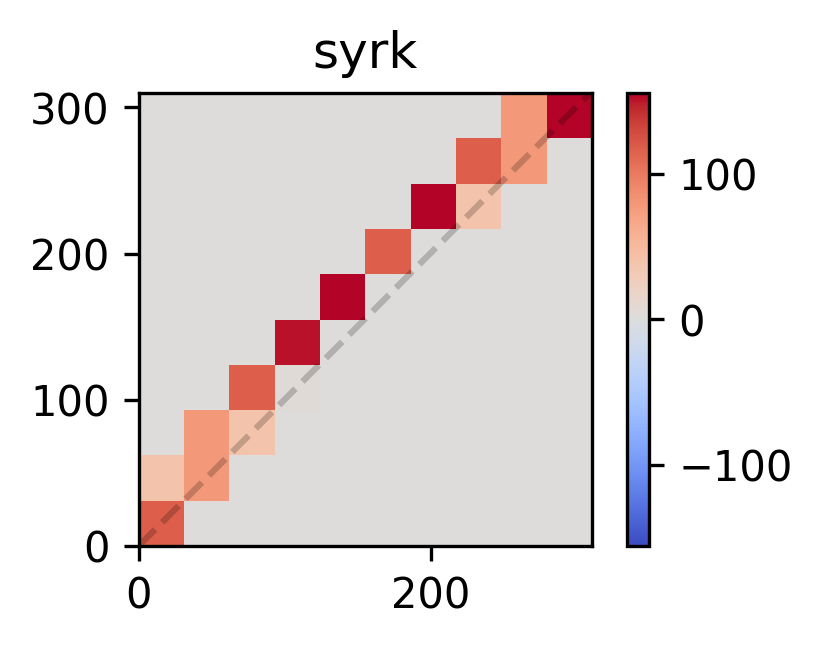}\hfill
    \includegraphics[width=.24\textwidth,height = .2\linewidth]{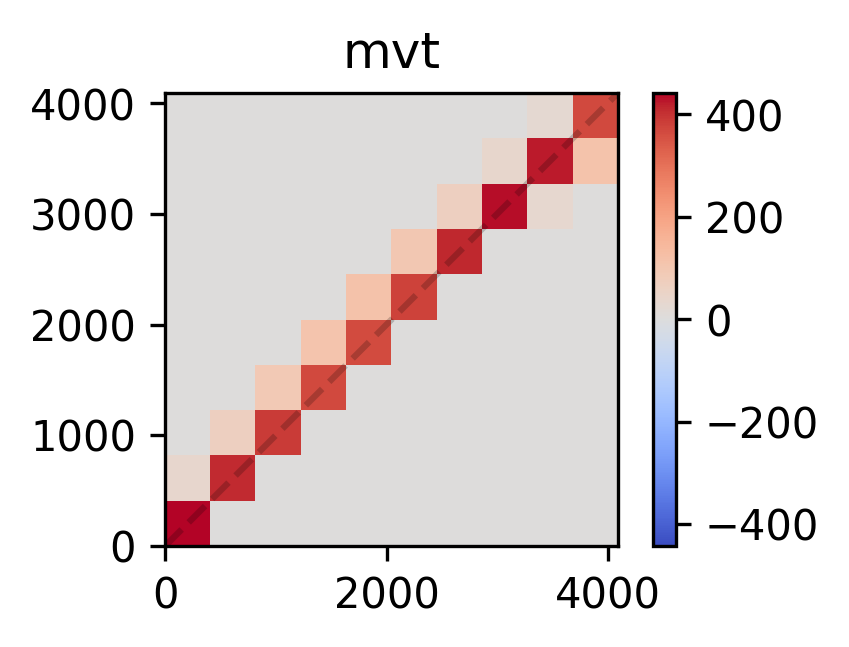}\hfill
    \includegraphics[width=.25\textwidth,height = .2\linewidth]{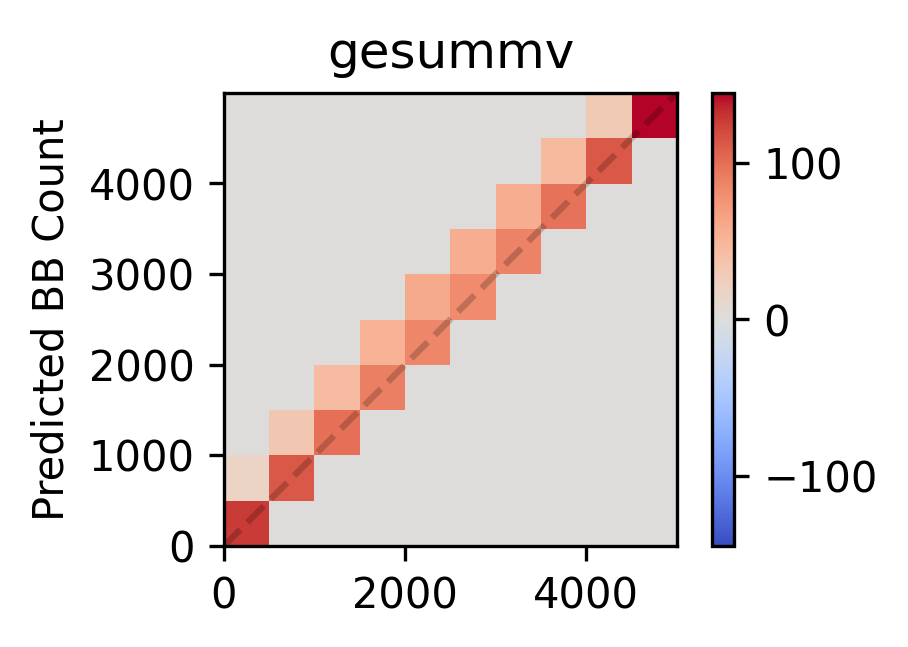}\hfill
    \includegraphics[width=.25\textwidth,height =.19\linewidth]{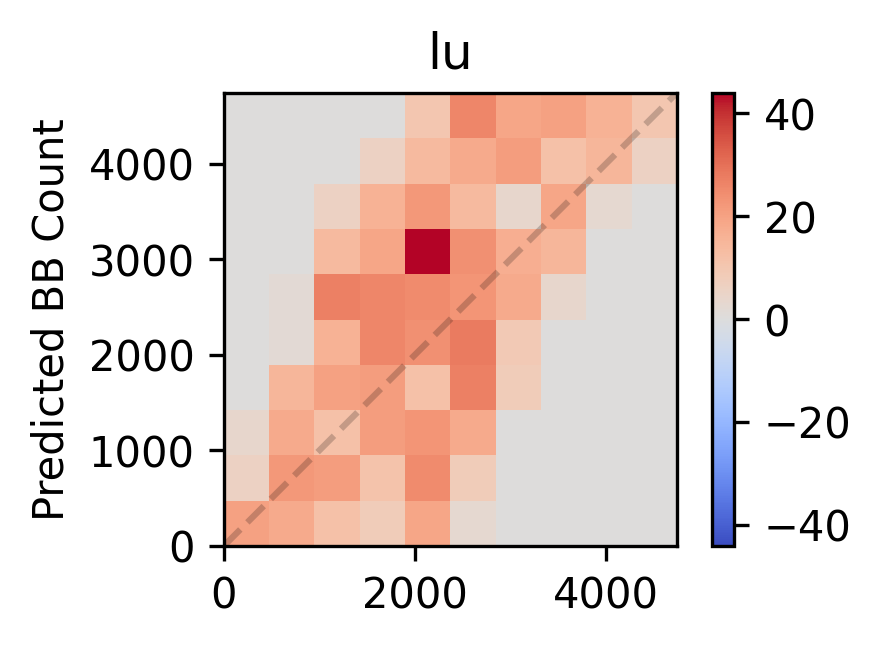}\hfill
    \includegraphics[width=.25\textwidth,height = .2\linewidth]{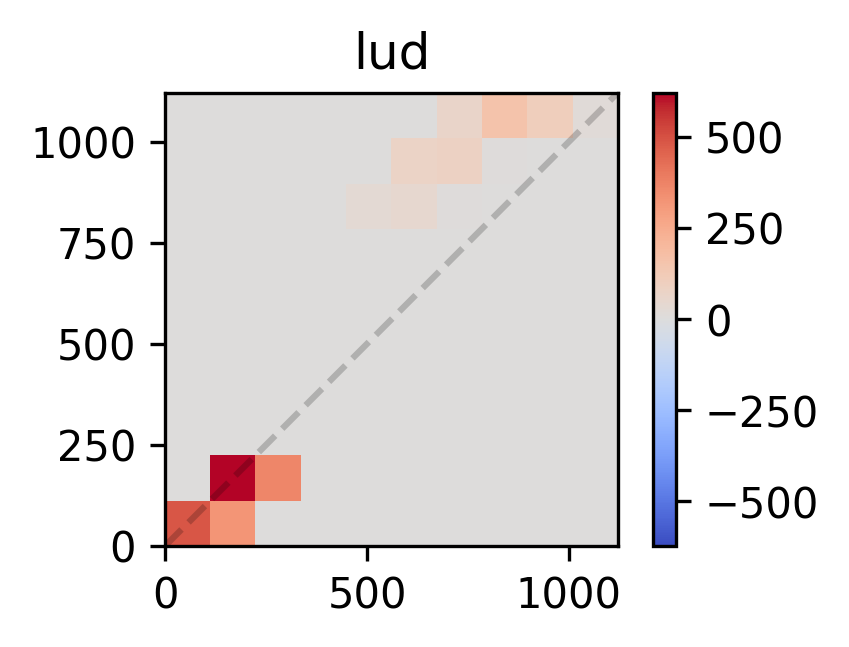}\hfill
    \includegraphics[width=.24\textwidth,height = .2\linewidth]{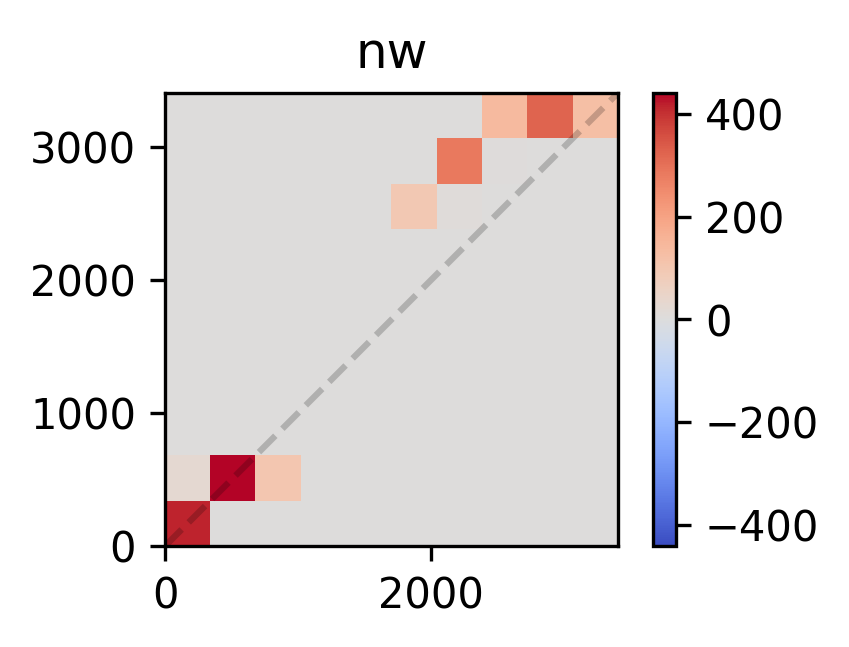}\hfill
    \includegraphics[width=.25\textwidth,height = .2\linewidth]{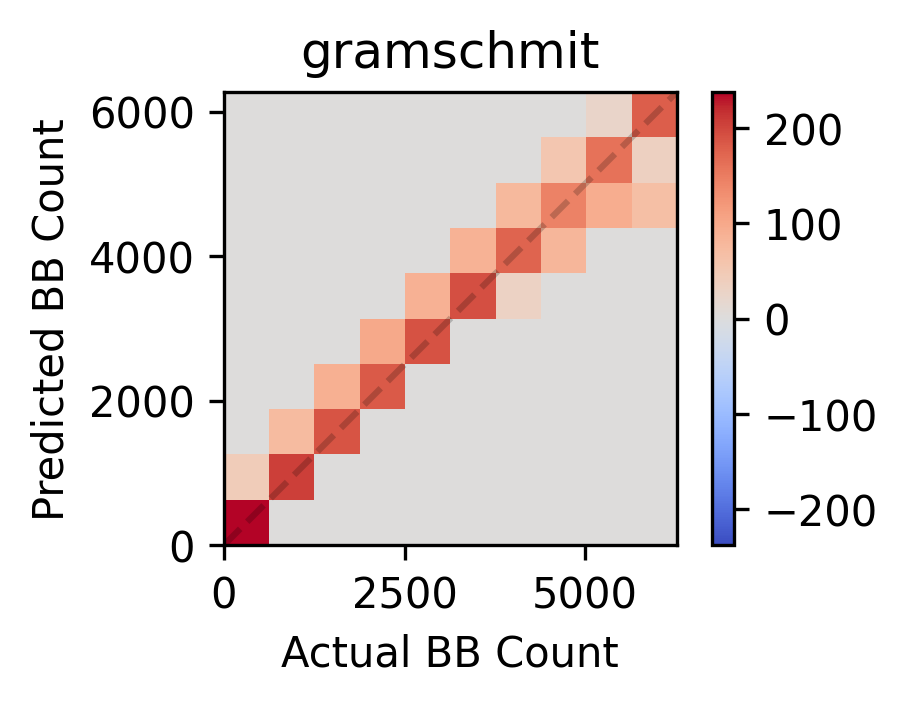}\hfill
    \includegraphics[width=.25\textwidth,height =.2\linewidth]{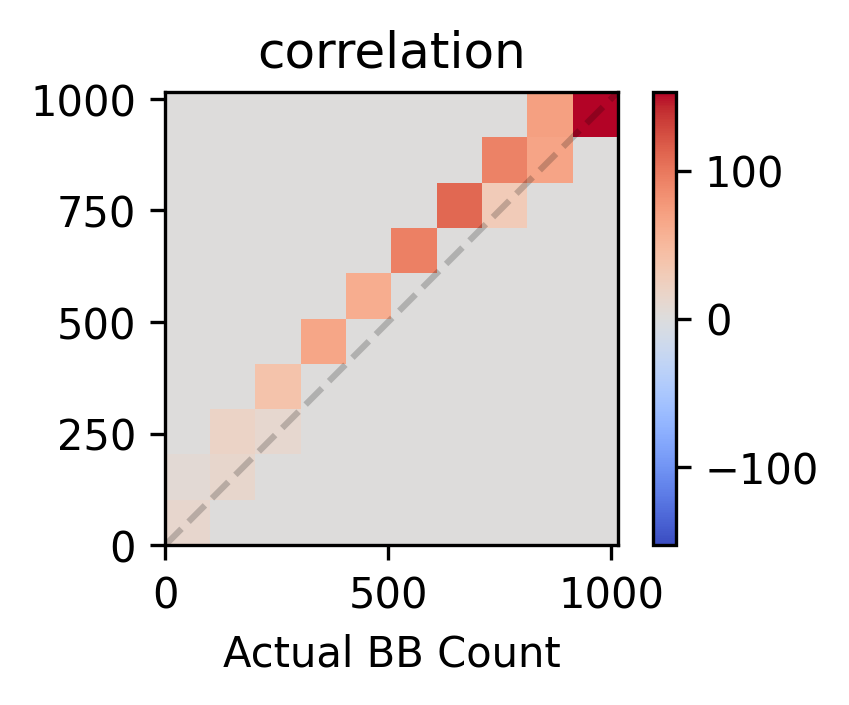}\hfill
    \includegraphics[width=.25\textwidth,height = .2\linewidth]{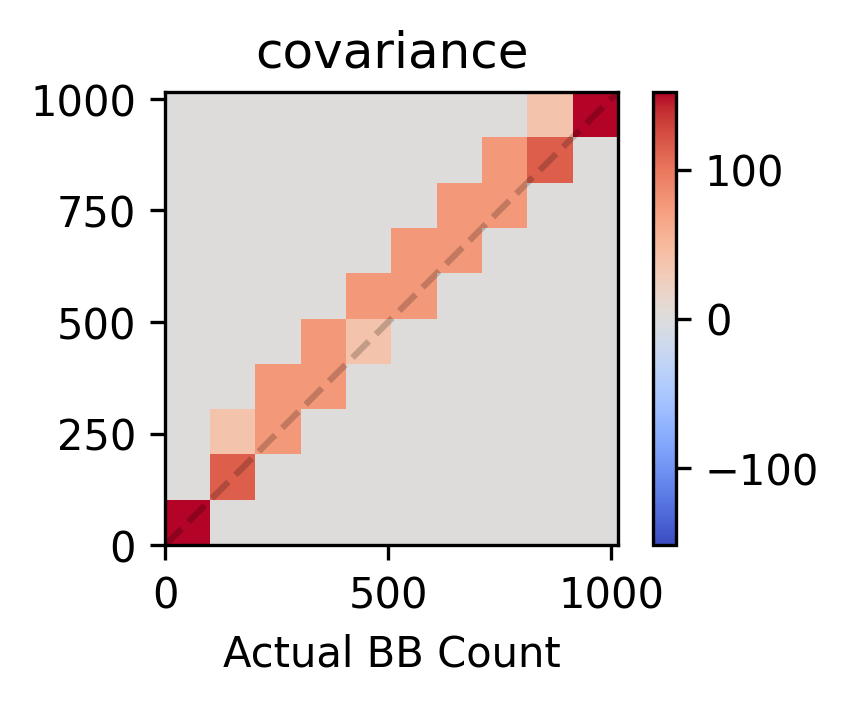}\hfill
    \includegraphics[width=.24\textwidth,height = .2\linewidth]{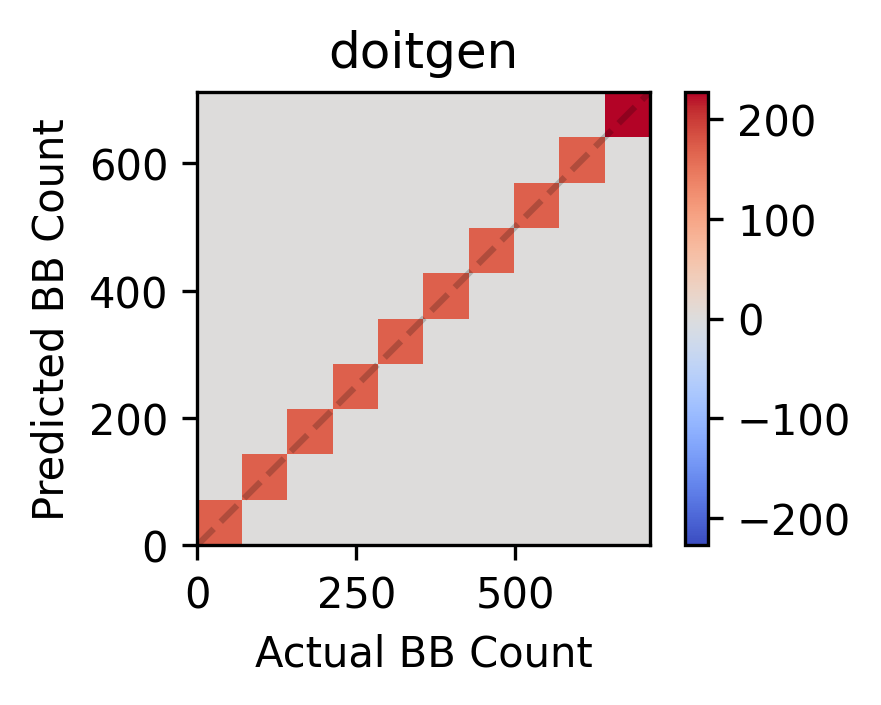}\hfill
   \vspace{-3mm}
    \caption{Heatmaps showing extrapolation of BB Count on high input values vs.\ actual basic block counts for PNN. From the plots above, we see that our model can accurately extrapolate for high input configurations while training on low inputs.}
    \label{Fig:heatmap}

\end{figure}
\section{Results}
\label{sec:exp}
\subsection{Experimental Setup}
We choose 16 broadly used GPU benchmarks listed in Table~\ref{dataset} to verify our prediction models. We use 12 applications from the Polybench benchmark suite~\cite{grauer2012auto} and four applications from the Rodinia benchmark suite~\cite{che2009rodinia}. We train the PNN model using these benchmarks with the different training and testing datasets described in Section~\ref{train-method}. 

\subsection{Predictions Results}
\label{sec:pred-result}
First, we evaluate the performance of our trained models using the test dataset from the random splitting of the sample set. We calculate the MSE with respect to the ground truth for each benchmark. Table~\ref{random-result} shows the average MSE and correlation statistics for predicting basic block counts for unseen test inputs from the random dataset with an overall MSE of less than $0.023$. For all benchmarks except for Gaussian, mvt, lud, and lu, the MSE for predicting basic block counts is close to zero (MSE $< 0.003$). Other applications have less than $0.012$ MSE error for random test set prediction except for the lu benchmark. We also find good Pearson and Spearman correlation values for almost all the benchmark applications (Table~\ref{random-result}).

To further investigate, we separate the low input values and high input values from the dataset by setting a cut-off value. We train the PNN model with samples that have only low input values for each benchmark and evaluate the model with large input instances. Table~\ref{exp-result} shows the extrapolation result of basic block counts on high input values when the model is trained using low input values as well as mixed input values. Among the 16 applications, gaussian, lu, nw, and lud have greater than $0.1$ MSE. Investigating these datasets shows that they have a significant number of zero basic block counts. Similar to random prediction results, we find higher correlation values which indicate a good linear relationship between extrapolation and actual basic block counts in the PNN model. 

The reason for obtaining correlation values less than 0.95 in certain applications, such as lu and Gaussian, is the multiple executions of the same kernel. Depending on the input size, these application kernels are invoked multiple times to apply the algorithm to varied sections of the entire matrix in each iteration. We observed that the basic block (BB) counts for the initial executions of each recurrent kernel, consistent with Nvidia Nsight methodology, remain constant. Yet, when aggregating the BB counts across all executions, the resulting data yields a less accurate model than other applications. We opted for the aggregate of all iterations as it offers a more comprehensive perspective on the overall application performance. The PNN model, on average, shows $93.6\%$ accuracy for extrapolation learned from lower input configuration and $97.7\%$ accuracy for random instance prediction.

A better representation of extrapolation is shown in Figure~\ref{Fig:heatmap} for PNN. The heatmaps show extrapolation of basic block execution counts on high input values with respect to the actual basic block counts for those inputs. Each heatmap includes all basic block predictions for one benchmark. The extrapolated BB counts are expected to be close to the actual count i.e. we expect all points along the identity line $(y=x)$. We find higher density near the identity line for most benchmarks except Gaussian, lu, lud, and nw. The heatmaps represent the closeness of our extrapolation counts with actual counts.



\begin{figure}[tb!]
	\centering
	\includegraphics[width=.51\linewidth]{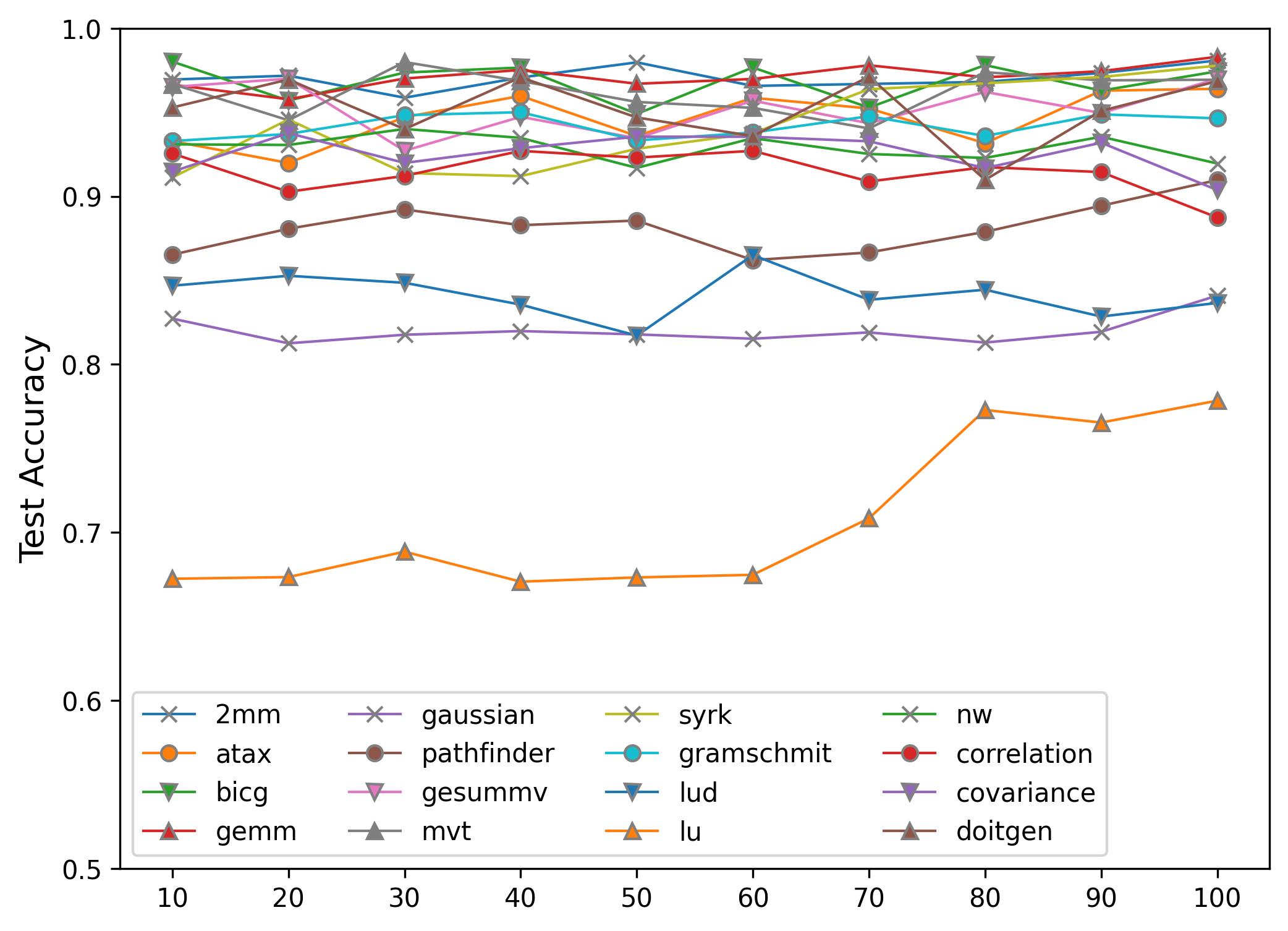}
    \includegraphics[width=.47\linewidth]{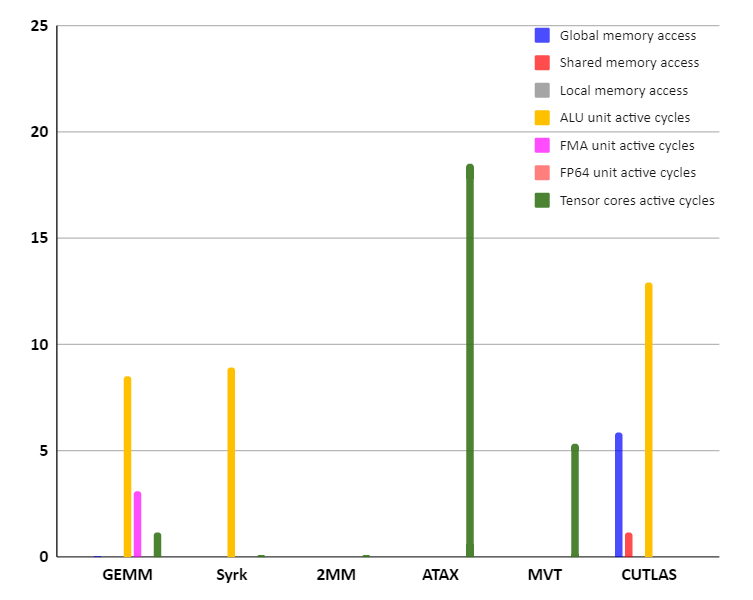}
    \caption{(left) Training dataset sample percentage plot against test data accuracy of all benchmarks for PNN model. (Right) Percentage error for all benchmarks and metrics} 
    \label{fig:sample-roc-pnn}
\end{figure}

Figure~\ref{fig:sample-roc-pnn}(left) shows the change in test accuracy when we vary the training sample percentage for the PNN model. This plot represents the relationship between model accuracy and training sample size. From the plot, high test accuracy for a small fraction of training samples indicates that we have sufficient samples for each benchmark application. We find that test accuracy decreases with higher samples for some benchmarks; they actually fit better with a small sample size. Although the accuracy for those applications shows slight improvement initially with increasing sample size, it decreases later with more samples. Overall, we find that PNN model is sensitive to the sample size; the test accuracy slightly increases for most benchmarks with increasing training sample size.

\subsection{BB Prediction: Case Study}

In our quest to elucidate the capabilities of our machine learning model, particularly in its fine-grained performance prediction at the Basic Block (BB) level, we turned our focus to CUDA benchmarks. We chose a wide range of benchmarks ranging from linear algebra applications such as 2mm, gemm, atax, and mvt to complex applications and ending with more complex machine learning benchmarks such as cutlass. This varied suite effectively underscores the adaptability of our model.

Figure~\ref{fig:sample-roc-pnn}(right) shows the Global memory access errors are consistently low across most benchmarks, but there is an exception in the last benchmark, which exhibits an average error of 5.88\%. For active cycles of the different units measured, we found that the average error is 2.3\% for the FMA and 10.66\% for the ALU, while the maximum average error which comes from the ATAX application is 18.5\%.

\section{Discussion \& Conclusion}
\label{sec:conclusion}
This work presents a Basic Block count prediction tool, BB-ML. BB-ML uses probabilistic and regression networks to predict GPU BB execution counts. 

Predicting the counts of BBs executions can be valuable for various purposes, such as performance analysis, code optimization, and enhancing program understanding. With an accurate prediction of BBs execution count, we do not need to perform profiling to analyze the runtime behavior of a program or the active cycles for the included computational units. The predicted BBs could provide a static overview of potential performance bottlenecks without requiring runtime execution. Additionally, we can collect memory traces to perform a comprehensive performance memory analysis in combination with the predicted basic blocks. This synthesis becomes particularly powerful when employing performance analysis tools such as PPT-GPU~\cite{arafa2019ppt} for a holistic evaluation of GPU applications.

In our research, we presented a probabilistic Poisson Deep Neural Network that has showcased remarkable proficiency in handling a broad range of input values with good BB prediction accuracy. Simultaneously, our performance prediction model has effectively demonstrated its capability in gauging the benchmarks' performance. The PNN model stand as testament to our commitment to pushing the boundaries in GPU application analysis and optimization.

\section{Acknowledgement}
The authors would like to thank the anonymous reviewers. Paper LAUR is LA-UR-23-29323.

\balance
\bibliographystyle{plainurl}
\bibliography{main-arxiv}

\end{document}